\documentclass[letterpaper]{article} 
\usepackage{aaai2026}  
\usepackage{times}  
\usepackage{helvet}  
\usepackage{courier}  
\usepackage[hyphens]{url}  
\usepackage{graphicx} 
\usepackage{natbib}  
\usepackage{caption} 
\usepackage{algorithm}
\usepackage{algorithmic}

\usepackage{newfloat}
\usepackage{listings}
\DeclareCaptionStyle{ruled}{labelfont=normalfont,labelsep=colon,strut=off} 
\floatstyle{ruled}
\newfloat{listing}{tb}{lst}{}
\floatname{listing}{Listing}
\title{First-Order Representation Languages for Goal-Conditioned RL}

\author{%
    Simon St\r{a}hlberg, Hector Geffner
    }
\affiliations{%
    RWTH Aachen University, Germany \\
    simon.stahlberg@gmail.com, hector.geffner@ml.rwth-aachen.de
}
\usepackage{bibentry}

\usepackage{etoolbox}
\AtBeginEnvironment{equation}{\footnotesize}
\AtBeginEnvironment{equation*}{\footnotesize}
\AtBeginEnvironment{align}{\footnotesize}
\AtBeginEnvironment{align*}{\footnotesize}
\AtBeginEnvironment{gather}{\footnotesize}
\AtBeginEnvironment{gather*}{\footnotesize}
\AtBeginEnvironment{multline}{\footnotesize}
\AtBeginEnvironment{multline*}{\footnotesize}

\usepackage[utf8]{inputenc}
\usepackage{amsfonts}
\usepackage{amsmath}
\usepackage{amssymb}
\usepackage{amsthm}
\usepackage{bm}
\usepackage{booktabs}
\usepackage{makecell}
\usepackage{subcaption}
\usepackage{tikz}
\usepackage{xcolor}

\usepackage{pgfplots}
\pgfplotsset{compat=1.18}
\usepgfplotslibrary{fillbetween}
\usetikzlibrary{positioning}
\usetikzlibrary{pgfplots.groupplots}

\newcommand{\predicates}{\ensuremath{\mathcal{P}}}
\newcommand{\predicate}{P}
\newcommand{\atom}{p}
\newcommand{\objects}{\ensuremath{\mathcal{O}}}
\newcommand{\object}{o}
\newcommand{\actionschemas}{\ensuremath{\hat{\mathcal{A}}}}
\newcommand{\actionschema}{\ensuremath{\hat{a}}}

\newcommand{\applicable}[1]{\ensuremath{\mathcal{A}[#1]}}
\newcommand{\action}[1]{\ensuremath{\textit{a}_{#1}}}
\newcommand{\actionname}{A}
\newcommand{\initialstate}{\ensuremath{\mathcal{I}}}
\newcommand{\goal}{\ensuremath{\mathcal{G}}}

\newcommand{\state}[1]{\ensuremath{\textit{s}_{#1}}}

\newcommand{\mlp}{\text{MLP}}

\newcommand{\rgnn}{\text{R-GNN}}

\newcommand{\emb}{\bm{f}}

\newcommand{\groundatomset}{\ensuremath{\mathcal{S}}}
\newcommand{\goalschemas}{\ensuremath{\hat{\bm{G}}}}
\newcommand{\goalschema}{\ensuremath{\hat{\goal}}}

\newcommand{\refine}{\textsc{Refine}}
\newcommand{\hindsightgoal}{\textsc{HindsightGoal}}

\newtheorem{definition}{Definition}

\nocopyright

\begin{document}

\maketitle

\begin{abstract}
    First-order relational languages have been used in MDP planning and reinforcement learning (RL) for two main purposes:
specifying MDPs in compact form, and representing and learning policies that are general and not tied to specific instances or state spaces.
In this work, we instead consider the use of first-order languages in goal-conditioned RL and generalized planning.
The question is how to learn goal-conditioned and general policies when the training instances are large and the goal cannot be reached by random exploration alone.
The technique of Hindsight Experience Replay (HER) provides an answer to this question: it relabels unsuccessful trajectories as successful ones by replacing the original goal with one that was actually achieved.
If the target policy must generalize across states and goals, trajectories that do not reach the original goal states can enable more data- and time-efficient learning.
In this work, we show that further performance gains can be achieved when states and goals are represented by sets of atoms.
We consider three versions: goals as full states, goals as subsets of the original goals, and goals as lifted versions of these subgoals.
The result is that the latter two successfully learn general policies on large planning instances with sparse rewards by automatically creating a curriculum of easier goals of increasing complexity.
The experiments illustrate the computational gains of these versions, their limitations, and opportunities for addressing them.

\end{abstract}

\section{Introduction}
Reinforcement learning (RL) provides a powerful framework for training agents to maximize rewards~\cite{sutton-barto-1998,bertsekas-tsitsiklis-1996}.
A significant challenge in RL arises in scenarios with large state spaces and sparse rewards, where random exploration often fails to achieve successful outcomes, leading to learning deadlock.
To mitigate this issue, \emph{Hindsight Experience Replay (HER)}~\cite{andrychowicz-et-al-nips2017} has been proposed as an effective technique.
HER operates in the \emph{goal-conditioned RL} (GCRL) setting, where policies are developed to generalize across both states and goals~\cite{andrychowicz-et-al-nips2017,nasiriany-et-al-neurips2019,eysenbach-et-al-iclr2019,eysenbach-et-al-neurips2022,chane-sane-et-al-icml2021}.
The core concept in HER involves relabeling unsuccessful trajectories: instead of treating a trajectory as a failure to reach the intended goal $g$, it is reinterpreted as successfully reaching the goal $g'$ that was actually achieved.
By breaking the deadlock caused by sparse rewards, HER significantly accelerates learning, even in challenging tasks like solving the Rubik's cube~\cite{agostinelli-et-al-nmi2019}.

Interestingly, GCRL shares similarities with \emph{generalized planning}~\cite{martin-geffner-ai2004,boutilier-et-al-ijcai2001,srivastava-et-al-aaai2008,hu-degiacomo-ijcai2011,jimenez-segovia-jonsson-ker2019}, which focuses on learning policies that generalize across different states and goals within a planning domain.
This problem has been explored through various approaches, including SAT-based~\cite{bonet-et-al-aaai2019,frances-et-al-aaai2021} and reinforcement learning approaches~\cite{stahlberg-et-al-kr2022,stahlberg-et-al-kr2023}.
In the RL context, generalized planning resemble GCRL but differs in using relational, first-order languages to represent both states and goals, akin to STRIPS~\cite{guestrin-et-al-ijcai2003,sanner-boutelier-aij2009}.
Furthermore, the scope of generalization is explicitly defined by the lifted description of the planning domain~\cite{haslum-et-al-2019}.

This paper introduces and evaluates three variants of HER adapted for the planning setting.
The first, \emph{state HER}, relabels goals as sets (conjunctions) of ground atoms representing full states.
The second, \emph{propositional HER}, limits relabeled goals to subsets of ground atoms found in the original goal.
The third, \emph{lifted HER}, uses lifted versions of propositional goals that maintain essential structural dependencies.
Experimental results demonstrate that the latter two variants are particularly effective, automatically constructing a curriculum by progressively identifying more challenging trajectories (e.g., larger goal sizes and states farther from the initial state).
Unlike previous RL approaches to generalized planning~\cite{stahlberg-et-al-kr2023}, which were limited to small instances requiring precomputation of optimal values $V^*(s)$ for uniform sampling, our approaches are scalable to billions of states without such precomputation.

The rest of this paper is organized as follows: we first review relevant background in classical planning and GCRL, then describe how states, actions, and goals are encoded for the relational graph neural network (\rgnn{}) architecture, the RL algorithm used, and the considered HER variants.
We present our experiments and discuss the results, with related work interspersed throughout.

\section{Background}

We present a brief overview of classical planning, GCRL, and \rgnn{}, which form the foundation of our approach.

\subsection{Classical Planning}
A planning instance is defined as a tuple $\langle \predicates{}, \actionschemas{}, \objects{}, \initialstate{}, \goal{} \rangle$.
In this paper, we use the hat notation, $\hat{X}$, to denote first-order (lifted) constructs, which may contain variables as well as constants.
When the hat is omitted, $X$ refers to the ground (propositional) version, where all variables have been replaced by constants.
We denote $\langle \predicates{}, \actionschemas{} \rangle$ as the \emph{domain} and $\langle \objects{}, \initialstate{}, \goal{} \rangle$ as the \emph{problem}, where:
\begin{itemize}
    \item $\predicates{}$ represents a set of \emph{predicate symbols}.
    For each symbol $\predicate{} \in \predicates{}$, there is an associated \emph{arity} $n$.
    An \emph{atom} is denoted as $\predicate{}(x_1, \dots, x_n)$; if no term $x_i$ is a variable (i.e., all terms are objects), then it is a \emph{ground atom}.

    \item $\actionschemas{}$ denotes a set of \emph{action schemas}.
    Each action schema $\actionschema{} = \actionname{}(X_1, \dots, X_k)$ is defined by a name $\actionname{}$ and a list of parameters $X_1, \dots, X_k$ (variables).
    An action schema consists of a \emph{precondition}, a set of atoms that must be true in the current state for the action to be applicable, and an \emph{effect}, a set of atoms that describe how the state is modified when the action is applied.

    \item $\objects{}$ denotes a set of \emph{objects} (constants).

    \item $\initialstate{}$ is the \emph{initial state} and $\goal{}$ is the \emph{goal}.
    Both are \emph{states}, which are sets of ground atoms over $\predicates{}$ and $\objects{}$.
    A state $\state{}$ is considered a \emph{goal state} if and only if $\goal{} \subseteq \state{}$.
\end{itemize}

A \emph{ground action} is an instantiation of an action schema $\actionschema{}$, where all variables are replaced by objects from $\objects{}$, e.g., $\action{} = \actionname{}(o_1, \dots, o_k)$.
The objective in planning is to find a sequence of ground actions that, when applied to the initial state $\initialstate{}$, leads to a state $\state{}$ satisfying $\goal{} \subseteq \state{}$.
Each ground action is only applicable if its precondition holds in the current state, and its effect deterministically updates the state.

\subsection{Goal-Conditioned Reinforcement Learning}

We adopt a GCRL framework tailored to the characteristics of classical planning.
In our formulation, we assume that: the actions are \emph{deterministic}, the environment is \emph{fully observable}, and there is a \emph{fixed initial state}.
Additionally, we define a reward function $r(\state{}, \action{}, \goal{})$, and a discount factor $\gamma \in [0, 1]$.

In GCRL, each episode begins with an initial state $\state{0}$ and a goal $\state{g}$.
In our setting, the initial state and the goal is always fixed to $\state{0} = \initialstate{}$ and $\state{g}=\goal{}$, respectively.
At each time step $t$, the agent selects an action $\action{t} = \pi(\state{t}, \goal{}), \quad \action{t} \in \applicable{\state{t}}$, receives an immediate reward $r_t = r(\state{t}, \action{t}, \goal{})$, and transitions to a new state $\state{t+1}$, which is obtained by applying $\action{t}$ in $\state{t}$.
Here, $\pi$ denotes the agent's \emph{policy}, which decides the action to take based on the current state and goal.
The objective of the agent is to maximize its return, defined as the discounted sum of future rewards $R_t = \sum_{i=t}^{\infty} \gamma^{i-t} r_i$.

In planning, the primary objective is to solve problems by reaching a goal state.
This is typically modeled in one of two ways: a reward is given only upon reaching a goal state, or a constant negative reward is incurred until the problem is solved.
We use the latter, setting $r(\state{}, \action{}, \goal{}) = -1$ (receiving a penalty always).
Goal states are terminal, meaning the episode ends once a goal is reached and no further rewards are accumulated.
In states with no applicable ground actions, a dummy action is provided that leaves the state unchanged.

The Q-value function $Q(\state{}, \action{}, \goal{})$ represents the expected return of taking action $\action{}$ in state $\state{}$ while pursuing the goal $\goal{}$.
A greedy policy selects the action with the highest return: $\pi_Q(\state{}, \goal{}) = \arg\max_{\action{} \in \applicable{\state{}}} Q(\state{}, \action{}, \goal{})$.
An \emph{optimal policy} $\pi^*$ satisfies $Q^{\pi^*}(\state{}, \action{}, \goal{}) \geq Q^{\pi}(\state{}, \action{}, \goal{})$ for all states $\state{}$, ground actions $\action{}$, goals $\goal{}$, and any policy $\pi$.
Notably, all optimal policies share the same Q-value function, which in turn satisfies the Bellman optimality equation:
\begin{equation}
Q^*(\state{}, \action{}, \goal{}) = r(\state{}, \action{}, \goal{}) + \gamma \max_{\action{}' \in \applicable{\state{}'}} Q^*(\state{}', \action{}', \goal{}),
\end{equation}
where $\state{}'$ denotes the state resulting from applying $\action{}$ in $\state{}$.
This equation forms the foundation for Q-learning.

\section{Learning General Policies}

We begin by describing the computation of Q-values for each applicable action in a given state relative to a goal.
Then, we describe Deep Q-Network (DQN) with HER.

\subsection{Q-value Predictions}

In classical planning, states are represented as sets of ground atoms, where sizes vary across states, and the number of applicable actions dynamically changes.
In this paper, we use a relational graph neural network (\rgnn{}) architecture~\cite{stahlberg-et-al-icaps2022} to learn Q-value functions in classical planning domains.
The \rgnn{} operates on relational structures and can handle variable-size inputs, making it suitable for classical planning.
The \rgnn{} architecture is a convenient choice as it can directly process sets of ground atoms.
Previous work using this architecture~\cite{stahlberg-et-al-icaps2022,stahlberg-et-al-kr2022,stahlberg-et-al-kr2023} focused on learning value functions, which involves generating successor states and computing their values.
In contrast, our approach learns a Q-value function that directly estimates the value of executing a specific action in a given state.

For brevity, we omit a detailed description of the \rgnn{} architecture here and refer the reader to~\cite{stahlberg-et-al-icaps2022}.
Conceptually, it functions as a black-box, taking sets $\groundatomset{}'$ and $\objects'{}$ as inputs and generating embeddings $\emb{}(\object{})$ for each object $\object{} \in \objects'{}$.
We simply need to define the input sets $\groundatomset{}'$ and $\objects'{}$ to encompass the state, applicable actions, and the goal.
Additionally, we design the input so that the Q-values for all applicable ground actions are computed in parallel.

It is crucial that the input encoding distinguishes between the state, applicable actions and the goal.
To achieve this, we introduce new predicate symbols and objects in the input.
The state uses original predicate symbols $\predicates{}$, applicable actions are represented by new predicate symbols $\predicate{}_{\actionname{}}$ for each action schema $\actionname{}$, and the goal by new predicate symbols $\predicate{}_\goal{}$ for each goal predicate $\predicate{} \in \predicates{}$.
Additionally, new objects $\object{}_{\action{}}$ are introduced for each applicable action $\action{}$.
The input $\groundatomset{}'$ and $\objects{}'$ are defined as $\groundatomset{}' = \groundatomset{}_{\state{}} \cup \groundatomset{}_{\applicable{\state{}}} \cup \groundatomset{}_{\goal{}}$, where:
\begin{itemize}
    \item $\groundatomset{}_{\state{}} = \state{}$ is the set of atoms true in the given state;
    \item $\groundatomset{}_{\applicable{\state{}}} = \{ \predicate{}_{\actionname{}}(\object{}_{\action{}}, \object{}_1, \dots, \object{}_n) : \action{} = \actionname{}(\object{}_1, \dots, \object{}_n) \in \applicable{\state{}} \}$, with $\predicate{}_{\actionname{}}$ and $\object{}_{\action{}}$ representing new predicate symbols and objects, respectively; and
    \item $\groundatomset{}_{\goal{}} = \{ \predicate{}_\goal{}(\object{}_1, \dots, \object{}_n) : \predicate{}(\object{}_1, \dots, \object{}_n) \in \goal{} \}$, with $\predicate{}_\goal{}$ representing a new predicate symbol.
\end{itemize}
The set of objects is $\objects{}' = \objects{} \cup \{ \object{}_{\action{}} : \action{} \in \applicable{\state{}} \}$, the union of objects in the state and new objects representing applicable actions.
Notably, the atom $\predicate_{\actionname{}}(o_a, \dots)$ for action $a$ includes terms of $a$, indirectly connecting $o_a$ to state and goal atoms.

Initialization of the \rgnn{} requires knowledge of all predicate symbols.
No new predicate symbol depends on a specific problem instance, ensuring a fixed set of new symbols for each domain.
Consequently, the model can be trained on one instance set and applied to others within the same domain.
The output of $\rgnn{}(\groundatomset{}', \objects{}')$ is a set of embeddings $\emb{}(\object{})$ for each object $\object{} \in \objects{}'$.
To compute the Q-value for a specific action, the embedding of the corresponding action object $\object{}_{\action{}}$ is used alongside a state summary.
The Q-value function is defined as:
\begin{equation}
    Q(\state{}, \action{}, \goal{}) = \mlp{}\left(\emb{}(\object{}_{\action{}}), \sum_{o \in \objects{}} \emb{}(o)\right) \,.
\end{equation}
Here, the sum encompasses embeddings of original state objects (excluding action objects), ensuring the Q-value is conditioned on both action and overall state context, where the goal is implicit.
Although using only $\emb{}(\object{}_{\action{}})$ is possible, it empirically leads to poorer generalization.

\subsection{Deep Q-Network for GCRL}

The presented Q-value function is differentiable and trainable using standard deep reinforcement learning methods.
In this section, we briefly outline the training process for the Q-value function using a variant of the DQN algorithm~\cite{mnih-et-al-nature2015}.
DQN involves two primary stages: (1) \emph{experience generation} by interacting with the environment, and (2) \emph{Q-value function training} using the gathered experience.
These stages are interleaved, allowing the agent to learn from its interactions while exploring the environment.

\subsubsection{Experience Generation}

The initial step involves collecting experience tuples $\langle \state{t}, \action{t}, r_t, \state{t+1}, \goal{} \rangle$ through policy execution in the environment.
These tuples are stored in a \emph{replay buffer}, a finite-memory structure holding recent experiences.
Here, individual transition steps are stored in the replay buffer rather than entire trajectories.
This decorrelation is safe in our context as classical planning adheres to the Markov property.
This step also entails an exploration-exploitation trade-off; we use Boltzmann exploration where the probability of choosing action $\action{}$ in state $\state{}$ with goal $\goal{}$ is:
\begin{equation}
    \pi(\action{} \mid \state{}, \goal{}) = \frac{\exp\left(Q(\state{}, \action{}, \goal{})/T\right)}{\sum_{\action{}' \in \applicable{\state{}}} \exp\left(Q(\state{}, \action{}', \goal{})/T\right)},
\end{equation}
with $T > 0$ regulating exploration versus exploitation, with higher $T$ increasing exploration.

\subsubsection{Optimization}

The subsequent step optimizes $Q(\state{t}, \action{t}, \goal{})$ via \emph{mini-batch gradient descent} on the loss:
\begin{gather}
    \mathcal{L} = (Q(\state{t}, \action{t}, \goal{}) - y_t)^2 \,, \\
    y_t = r_t + \gamma \max_{\action{}' \in \applicable{\state{t+1}}} Q(\state{t+1}, \action{}', \goal{}) \,,
\end{gather}
where $y_t$ is the target value and $\gamma$ is the discount factor.
The value of $y_t$ is computed using a target network that is updated every $k$ episodes using the weights of the main network, with $k$ set to 1 in our experiments.
The mini-batch is sampled from the replay buffer.

\subsubsection{Experience Refinement}

Though the prior steps suffice for training a Q-value function, experience quality can be improved.
Hindsight Experience Replay (HER) is one approach, generating extra experience by reinterpreting trajectories with different goals~\cite{andrychowicz-et-al-nips2017}.
Here, the refinement step is a function $\refine{}$ operating on a trajectory $\tau$ and a problem, yielding refined trajectories $\tau_1, \dots, \tau_m$.
The input includes the problem for problem-specific strategies, like those discussed later.
The refined trajectories are stored in the replay buffer.

\section{Hindsight Experience Replay Variations}
\label{section:her}

Hindsight Experience Replay (HER) is a technique designed to improve sample efficiency in environments with sparse rewards.
The core idea is to reinterpret unsuccessful episodes by substituting the original goal with an alternative goal that was achieved during the episode.
This allows the agent to learn from every experience, regardless of whether the original goal was met.
More precisely, an experience tuple is stored as $\langle \state{}, \action{}, r, \state{}', \goal{}\rangle$, where $\goal{}$ is the original goal.
HER then generates additional transitions by:
sampling a new goal $\goal{}'$ from states encountered later in the same episode;
recalculating the reward using $\goal{}'$; and
augmenting the replay buffer with these hindsight experiences.
This procedure is implemented by the aforementioned $\refine{}$ function.

We introduce three variants of HER tailored to the classical planning setting: \textit{state HER}, \textit{propositional HER}, and \textit{lifted HER}.
State HER closely mirrors the original HER algorithm, whereas propositional and lifted HER exploit the structure of the goal to generate more informative hindsight experiences.
The only difference between the three variants is the definition of the \hindsightgoal{} function, which determines the hindsight goal for each subtrajectory.
In all three cases, the algorithm greedily extracts the longest possible non-overlapping, cycle-free subtrajectories that achieve the hindsight goal only at their final states.
Although it is possible to allow overlapping subtrajectories, this may reduce the diversity of experiences in the replay buffer.

\subsection{State HER}

In standard HER~\cite{andrychowicz-et-al-nips2017}, the hindsight goal is defined as the state achieved at the end of the subtrajectory.
Formally,
\begin{equation}
    \hindsightgoal{}(\state{}, \goal{}) = \state{}.
\end{equation}
This approach treats the entire final state, including both fluent and static atoms, as the new goal.
While static atoms are always satisfied and could be omitted for conciseness, we retain them here to remain consistent with the original HER formulation.

\subsection{Propositional HER}

Propositional HER differs from state HER by exploiting the structure of the original goal rather than ignoring it entirely.
Instead of substituting the goal with the final state, propositional HER constructs a new goal by selecting the largest subset of the original goal that is achieved at the end of the subtrajectory.
For instance, in the Blocks domain, if the original goal is to build a tower of blocks but only a partial stack is achieved, the hindsight goal becomes the maximal subset of goal atoms satisfied in the final state.
Formally, the hindsight goal is defined as:
\begin{equation}
    \hindsightgoal{}(\state{}, \goal{}) = \arg\max_{\goal{}' \subseteq \goal{}} \left\{ |\goal{}'| : \goal{}' \subseteq \state{} \right\}.
\end{equation}
This method is especially useful in domains where goals are expressed as conjunctions of atoms, letting the agent to learn from partial achievements.
However, in cases where the goal consists of a single atom, propositional HER offers no advantage over the original goal, as no proper subset exists.

\subsection{Lifted HER}

The limitations of propositional HER can be addressed by lifting goals to the first-order level, enabling the generation of \emph{analogous} hindsight goals that capture structural similarities rather than requiring exact matches.
For example, in the Blocks domain, a trajectory may construct a tower using blocks different from those specified in the original goal.
Such a trajectory is still valuable, as it demonstrates the ability to achieve the underlying goal structure.
To achieve this, lifted HER lifts the original goal to a first-order representation by replacing constants with variables, allowing for alternative groundings.
However, this process requires making implicit constraints explicit, such as ensuring variables are assigned to distinct objects and that the goal forms a similar structure (e.g., a tower).

\begin{algorithm}[t]
    \caption{Generation of lifted goals.}
    \begin{minipage}[t]{1.0\linewidth}
        \footnotesize
        \begin{algorithmic}[1]
            \STATE \textbf{Input:} Grounded goal $\goal{}$
            \STATE \textbf{Output:} Lifted goals $\goalschemas{}$
            \STATE Initialize $\goalschemas{} \gets \emptyset$
            \STATE Construct a goal-dependency graph $G(\goal{}) = (V, E)$:
            \STATE Identify connected components $C_1, \dots, C_k$ of $G$
            \STATE Let $C_i^*$ be the set of all subgraphs of $C_i$ containing at most one connected component
            \STATE Compute the cartesian product $C^* = C_1^* \times \dots \times C_k^*$, and combine subgraphs
            \FOR{each $(V', E') \in C^*$, $V' \not = \emptyset$}
                \STATE Let $\goalschema{}$ be $V'$ with objects replaced by variables
                \STATE For each pair of variables in $\goalschema{}$, add $X \not = Y$ to $\goalschema{}$
                \STATE Add $\goalschema{}$ to $\goalschemas{}$
            \ENDFOR
        \end{algorithmic}
    \end{minipage}
    \label{alg:lifted-goals}
\end{algorithm}

The first step is to construct a goal-dependency graph for $\goal{}$, which captures the relationships between atoms in the grounded goal based on shared objects.
For example, consider a goal that requires building a single tower in the Blocks domain.
We want to avoid generating lifted subgoals such as $\textsc{On}(X, Y) \wedge \textsc{On}(Z, W)$, where all four variables can be bound to different objects.
In practice, such subgoals are likely to correspond to disconnected structures, which are not meaningful in the context of the original goal.
The goal-dependency graph prevents the enumeration of these structurally irrelevant subgoals by ensuring that only connected subgraphs, corresponding to coherent goal structures, are considered during lifted goal generation.
\begin{definition}
    A \emph{goal-dependency graph} $G(\goal{}) = (V, E)$ is an undirected graph where each vertex $v \in V$ corresponds to an atom in $\goal{}$, and an edge $(\atom{}, \atom{}') \in E$ exists if and only if $\atom{}$ and $\atom{}'$ share at least one object in its arguments.
\end{definition}

The second step is to identify subgraphs of $G(\goal{})$ that do not split any \emph{existing} connected components into multiple components.
In the Blocks example, this means that the subgraph $\textsc{On}(a, b) \wedge \textsc{On}(b, c)$ is valid, while $\textsc{On}(a, b) \wedge \textsc{On}(c, d)$ is not as it splits the goal into two disconnected components.
Note that $G(\goal{})$ may contain multiple connected components, we generate valid subgoals for each connected component separately and later combine them.

The third step is to lift the propositional subgoals to first-order representations.
This is done by replacing the constants in the subgraph with variables, and adding inequality constraints to ensure that the variables are distinct.
For example, the propositional subgoal $\textsc{On}(a, b) \wedge \textsc{On}(b, c)$ is lifted to $\textsc{On}(X, Y) \wedge \textsc{On}(Y, Z) \wedge (X \neq Y) \wedge (Y \neq Z)$.

Algorithm~\ref{alg:lifted-goals} outlines the procedure for generating lifted goals.
Given a grounded goal $\goal{}$, the algorithm produces a set of lifted goal schemas $\goalschemas{}$.
We define $\textsc{Ground}(\goalschema{}, \state{})$ as a grounding of the goal schema $\goalschema{}$ that is satisfied in the state $\state{}$, or $\bot$ if no such grounding exists.
In other words, this operation returns a propositional instantiation of $\goalschema{}$ that holds in $\state{}$.
The \hindsightgoal{} function for lifted HER is then defined as follows:
\begin{gather}
    \goalschema{}^* = \arg\max_{\goalschema{} \in \goalschemas{}} \left\{ |\goalschema{}| : \textsc{Ground}(\goalschema{}, \state{}) \neq \bot \right\} \\
    \hindsightgoal{}(\state{}, \goal{}) = \textsc{Ground}(\goalschema{}^*, \state{})
\end{gather}
If there is no grounding, then $\hindsightgoal{}(\state{}, \goal{}) = \bot$.

\subsection{Example}

Consider a goal in the Blocks domain that requires stacking blocks $b_1$, $b_2$, $b_3$, and $b_4$ in the specific order, with $b_1$ on top and $b_4$ at the bottom.
Propositional HER would generate the following subgoals:
{%
    \begin{itemize}
        \item $\textsc{On}(b_1, b_2) \wedge \textsc{On}(b_2, b_3) \wedge \textsc{On}(b_3, b_4)$,
        \item $\textsc{On}(b_1, b_2) \wedge \textsc{On}(b_2, b_3)$, $\textsc{On}(b_2, b_3) \wedge \textsc{On}(b_3, b_4)$,
        \item $\textsc{On}(b_1, b_2)$, $\textsc{On}(b_2, b_3)$, $\textsc{On}(b_3, b_4)$.
    \end{itemize}
}

In contrast, lifted HER generates subgoal schemas that generalize beyond the specific blocks $b_1$, $b_2$, $b_3$, and $b_4$.
The resulting lifted subgoals are:
{%
    \begin{itemize}
        \item $\textsc{On}(X, Y) \wedge \textsc{On}(Y, Z) \wedge \textsc{On}(Z, W) \wedge X \not = Y \wedge \dots$,
        \item $\textsc{On}(X, Y) \wedge \textsc{On}(Y, Z) \wedge X \not = Y \wedge \dots$,
        \item $\textsc{On}(X, Y) \wedge X \not = Y$.
    \end{itemize}
}

In state HER, the hindsight goal is the entire final state of the subtrajectory, including atoms such as $\textsc{Clear}(b_1)$ and $\textsc{On-Table}(b_4)$ that are unrelated to the original goal.
Including these extra atoms might result in a goal that is harder to achieve than the original, since all blocks must be arranged in the specified configuration.

\section{Experiments}

\begin{table}[ht]
    \footnotesize
    \centering

\begin{tabular}{@{}lccc@{}}
\toprule
Domain        & Train                                & Validation                             & Test                                   \\ \midrule
Blocks        & 2 -- 14 blk                          & 15 -- 19 blk                           & 20 -- 50 blk                           \\
Childsnack    & 1 -- 6 chd                           & 7 -- 12 chd                            & 13 -- 50 chd                           \\
Childsnack-AF & 1 -- 10 chd                          & 11 -- 19 chd                           & 20 -- 119 chd                          \\
Delivery      & \makecell{2 -- 5 grd,\\1 -- 5 pkg}   & \makecell{6 -- 9 grd,\\1 -- 7 pkg}     & \makecell{10 -- 25 grd,\\6 -- 10 pkg}  \\
Gripper       & 1 -- 19 bal                          & 20 -- 29 bal                           & 30 -- 129 bal                          \\
Hiking        & \makecell{1 -- 6 cpl,\\1 -- 4 loc}   & \makecell{6 -- 10 cpl,\\2 -- 5 loc}    & \makecell{10 -- 26 cpl,\\5 -- 7 loc}   \\
Miconic       & \makecell{1 -- 19 ppl,\\2 -- 20 flr} & \makecell{20 -- 29 ppl,\\21 -- 30 flr} & \makecell{30 -- 80 ppl,\\31 -- 81 flr} \\
Reward        & \makecell{2 -- 12 grd,\\1 -- 7 rwd}  & \makecell{13 -- 15 grd,\\8 -- 13 rwd}  & \makecell{16 -- 29 grd,\\8 -- 14 rwd}  \\
Spanner       & 1 -- 14 nut                          & 15 -- 24 nut                           & 25 -- 50 nut                           \\
Visitall      & 2 -- 12 grd                          & 13 -- 15 grd                           & 16 -- 30 grd                           \\ \bottomrule
\end{tabular}

    \caption{%
        Overview of the training, validation, and test sets for each domain.
        The training set is used to fit the models, the validation set to select hyperparameters and the best checkpoint, and the test set to assess final performance.
        Instances are generated using one or two scaling parameters with even distribution.
        The parameters are as follows: balls (bal), children (chd), couples (cpl), floors (flr), grid size (grd), locations (loc), nuts (nut), packages (pkg), people (ppl), and rewards (rwd).
        The suffix "AF" denotes allergy-free instances (i.e., without gluten-intolerant children).
        }
    \label{tab:data}
\end{table}

\begin{table}[ht]
    \footnotesize
    \centering
    \begin{tabular}{lccccc}
\toprule
 &  & \multicolumn{4}{c}{LAMA} \\
\cmidrule(lr){3-6}
 &  &  & \multicolumn{3}{c}{Plan Length} \\
\cmidrule(lr){4-6}
Domain & \# & Cov. & Total & Median & Mean \\
\midrule
Blocks & 100 & 100 & 23250 & 218.0 & 232.5 \\
Childsnack & 100 & 24 & 2361 & 89.0 & 98.4 \\
Childsnack-AF & 100 & 65 & 16965 & 261 & 261.0 \\
Delivery & 100 & 99 & 27337 & 276 & 276.1 \\
Gripper & 100 & 100 & 23800 & 238.0 & 238.0 \\
Hiking & 100 & 26 & 2038 & 77.5 & 78.4 \\
Miconic & 100 & 100 & 19556 & 195.5 & 195.6 \\
Reward & 100 & 99 & 12054 & 116 & 121.8 \\
Spanner & 100 & 0 & 0 & 0 & 0.0 \\
Visitall & 100 & 100 & 53004 & 475.0 & 530.0 \\
\midrule
Total & 1500 & 1212 & 312029 & 213.5 & 257.4 \\
\bottomrule
\end{tabular}

    \caption{%
        The coverage and the plan length (total, median, and mean) for each domain for LAMA.
        The plan length is for the first solution found. 
        LAMA solved 80.8\% of the instances.
    }
    \label{table:coverage:lama}
\end{table}

\begin{table*}[ht]
    \footnotesize
    \centering
    \begin{tabular}{lcccccccccccc}
\toprule
 &  \multicolumn{4}{c}{Lifted HER} & \multicolumn{4}{c}{Propositional HER} & \multicolumn{4}{c}{State HER} \\
\cmidrule(lr){2-5} \cmidrule(lr){6-9} \cmidrule(lr){10-13}
 &  & \multicolumn{3}{c}{Plan Length} &  & \multicolumn{3}{c}{Plan Length} &  & \multicolumn{3}{c}{Plan Length} \\
\cmidrule(lr){3-5} \cmidrule(lr){7-9} \cmidrule(lr){11-13}
Domain & Cov. & Total & Median & Mean & Cov. & Total & Median & Mean & Cov. & Total & Median & Mean \\
\midrule
Blocks & \textbf{100} & 9672 & 98.0 & 96.7 & \textbf{100} & 10730 & 100.0 & 107.3 & 98 & 10876 & 112.0 & 111.0 \\
Blocks-L & \textbf{100} & 9692 & 97.0 & 96.9 & \textbf{100} & 9802 & 98.0 & 98.0 & 78 & 11456 & 95.0 & 146.9 \\
Childsnack & 56 & 4358 & 69.0 & 77.8 & \textbf{100} & 9263 & 90.5 & 92.6 & 59 & 6309 & 111 & 106.9 \\
Childsnack-AF & \textbf{100} & 23424 & 211.5 & 234.2 & \textbf{100} & 21150 & 211.5 & 211.5 & 53 & 8240 & 141 & 155.5 \\
Delivery & \textbf{12} & 3506 & 243.5 & 292.2 & 11 & 1837 & 137 & 167.0 & 0 & 0 & 0 & 0.0 \\
Gripper & \textbf{100} & 23800 & 238.0 & 238.0 & \textbf{100} & 23988 & 238.0 & 239.9 & \textbf{100} & 24400 & 238.0 & 244.0 \\
Gripper-L & \textbf{100} & 23800 & 238.0 & 238.0 & \textbf{100} & 23858 & 239.0 & 238.6 & \textbf{100} & 23950 & 238.0 & 239.5 \\
Hiking & 0 & 0 & 0 & 0.0 & 0 & 0 & 0 & 0.0 & 0 & 0 & 0 & 0.0 \\
Miconic & \textbf{100} & 15999 & 159.0 & 160.0 & \textbf{100} & 15837 & 158.5 & 158.4 & \textbf{100} & 17955 & 180.5 & 179.6 \\
Miconic-L & \textbf{100} & 15870 & 158.5 & 158.7 & \textbf{100} & 15929 & 159.0 & 159.3 & 0 & 0 & 0 & 0.0 \\
Reward & 58 & 5564 & 86.5 & 95.9 & \textbf{72} & 6596 & 87.5 & 91.6 & 0 & 0 & 0 & 0.0 \\
Reward-L & \textbf{85} & 7941 & 89 & 93.4 & 78 & 7139 & 89.0 & 91.5 & 26 & 2566 & 95.5 & 98.7 \\
Spanner & \textbf{100} & 9555 & 94.0 & 95.5 & \textbf{100} & 9555 & 94.0 & 95.5 & 74 & 6766 & 90.0 & 91.4 \\
Visitall & \textbf{100} & 45329 & 409.0 & 453.3 & 88 & 34546 & 363.0 & 392.6 & 49 & 32018 & 654 & 653.4 \\
Visitall-L & 68 & 26247 & 340.0 & 386.0 & \textbf{87} & 36119 & 368 & 415.2 & 28 & 13706 & 483.0 & 489.5 \\
\midrule
Total & 1179 & 224757 & 136 & 190.6 & \textbf{1236} & 226349 & 136.5 & 183.1 & 765 & 158242 & 148 & 206.9 \\
\bottomrule
\end{tabular}

    \caption{%
        The number of solved instances and the plan length (total, median, and mean) for each domain for the learned models.
        There are 100 test instances per domain and 15 domains, resulting in a total of 1500 instances.
        See Table~\ref{tab:data} for the training, validation, and test sets.
        The suffix "L" indicates that training instances too large to be fully expanded were excluded.
        When generating a solution, a maximum of 1000 steps was allowed, and actions leading to previously visited states were explicitly excluded.
        Models trained with lifted HER solve 78.6\% of instances, while those trained with propositional HER solve 82.4\%.
        In contrast, models trained with state HER solve only 51.0\% of instances.
        The quality of the solutions can be compared to the solutions produced by LAMA (Table~\ref{table:coverage:lama}).
        }
    \label{tab:coverage}
\end{table*}

\begin{figure}[ht]
    \footnotesize
    \input{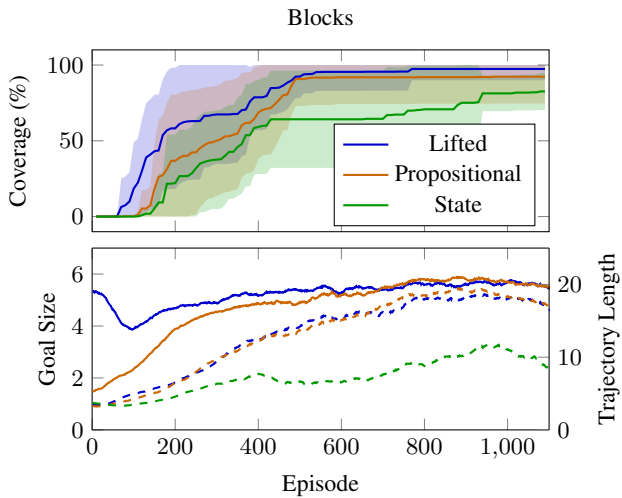}
    \caption{%
        The x-axis shows training episodes.
        \textbf{Top:} Solid line is mean test coverage of the best model up to each episode (cycle avoidance not used; episodes terminate on cycles). Shaded area is standard deviation.
        \textbf{Bottom:} Solid line (left y-axis) is mean goal size; dashed line (right y-axis) is mean trajectory length.
        Both plots use 10 seeds.
    }
    \label{fig:plots:blocks}
\end{figure}

\begin{figure}[ht]
    \footnotesize
    \input{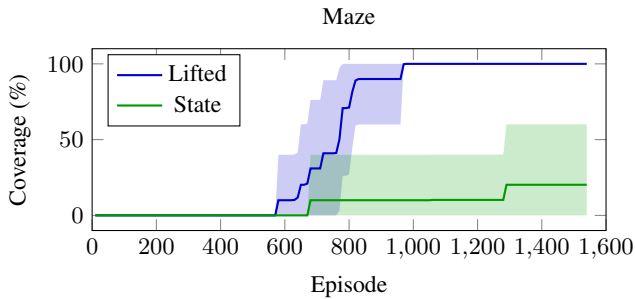}
    \caption{%
        See Figure~\ref{fig:plots:blocks} for a description of the plot.
        Prop. HER is not shown, as not a single trajectory was relabeled.
    }
    \label{fig:coverage:maze}
\end{figure}

We first describe the benchmark used in the experiments, then explain how the training and testing were set up.
Finally, we discuss the results of the experiments.

\subsection{Domains}

The domains used in the experiments are those from~\citeauthor{drexler-et-al-kr2024}~(\citeyear{drexler-et-al-kr2024}), where $C_2$ expressiveness is sufficient to distinguish all non-isomorphic states.
Models based on \rgnn{} architecture cannot fit the training data if there are value conflicts; however, the absence of value conflicts does not guarantee that a general policy can be learned.
To conserve space, we refer the reader to~\citeauthor{drexler-et-al-kr2024}~(\citeyear{drexler-et-al-kr2024}) for the domain descriptions.
The datasets used are shown in Table~\ref{tab:data}.

In addition, we also evaluate on a \textbf{Maze} domain.
In this domain, the task is to navigate a grid to reach a specific location.
This domain illustrates the limitations of propositional HER, as it fails to relabel trajectories in this domain.

\subsection{Setup}

We now describe the experimental setup for training and evaluation.
This includes details of the architecture, hyperparameters, and the training and testing procedures.
The methods were implemented using PyTorch, and used a lifted successor generator to ground goals~\cite{stahlberg-ecai2023}.\footnote{Code and data: \url{https://zenodo.org/records/17605619}}

\subsubsection{Model}

In contrast to the original \textsc{RNN}~\cite{stahlberg-et-al-icaps2022}, we use residual updates and layer normalization.
Furthermore, we use a readout from a random layer in addition to the readout from the last layer~\cite{bansal-et-al-neurips2022}, imposing the same loss on both.
This was necessary to train networks with $100$ layers.

\subsubsection{Hyperparameters}

The embedding size was set to $32$, using hard maximum for aggregation.
The learning rate started at $10^{-3}$ and decayed linearly to $10^{-6}$ over $300$ episodes.
The Boltzmann temperature decayed from $1.0$ to $0.1$ linearly over $600$ episodes.
The discount factor was $0.999$.
Each episode included $32$ optimization steps with a batch size of $32$.
Huber loss~\cite{huber-1964} (delta = $1.0$) was used instead of MSE.
The replay buffer size was $1000$, with prioritized experience replay~\cite{schaul-et-al-iclr2016} (priority exponent $0.6$, priority weight $0.4$).
We generated $4$ trajectories per episode, each up to $100$ steps.
The target network was updated after each episode using the main network’s weights.

\subsubsection{Training}

We trained $10$ models per method and domain, each with a unique random seed, to compute mean and standard deviation of coverage.
Training was performed on a system with an Intel Xeon Platinum 8352M CPU, $16$~GB RAM, and an NVIDIA A10 GPU with $24$~GB VRAM, lasting up to $12$ hours per model.
Model selection was based on validation coverage; in case of ties, shortest total solution length was used, and then randomly selected.

\subsubsection{Testing}

We evaluated the learned Q-value functions using a greedy policy:
at each step, the action with highest Q-value was selected, i.e., $\action{}^* = \arg\max_{\action{} \in \applicable{\state{}}} Q(\state{}, \action{}, \goal{})$.
The action $\action{}^*$ updated the state $\state{}$, repeated until $\goal{} \subseteq \state{}$ or after $1000$ steps.
Actions leading to previously visited states were excluded to avoid cycles.

\subsection{Baselines}

Learning general policies for planning domains with reinforcement learning (RL) is difficult because rewards are extremely sparse: agents usually receive no reward until the goal is reached, so random exploration almost never produces successful episodes and learning stalls.
Consequently, standard RL methods such as REINFORCE~\cite{williams-1992} and Actor-Critic~\cite{sutton-et-al-1999} generally fail to learn useful policies without a carefully designed curriculum.

To address reward sparsity, \citeauthor{gehring-et-al-icaps2022}~(\citeyear{gehring-et-al-icaps2022}) propose domain-independent heuristic reward shaping.
This technique is orthogonal to our approach and could be combined with it; we therefore omit it from our baselines.

Related work by \citeauthor{stahlberg-et-al-kr2023}~(\citeyear{stahlberg-et-al-kr2023}) learns general policies with RL but is limited to very small state spaces and does not scale to our larger instances.
Their method also requires sampling arbitrary initial states, whereas we always start from a fixed initial state.

Our primary baseline is \emph{state HER}, an adaptation of HER~\cite{andrychowicz-et-al-nips2017} to planning.
We compare our lifted and propositional HER variants against this baseline, and we evaluate the resulting plan quality against that of the LAMA planner~\cite{richter-et-al-ipc2011}.

\subsection{Results}

The main results are presented in Table~\ref{tab:coverage}, which reports the number of solved instances and plan lengths (total, median, and mean) for each domain.
Lifted HER solves $78.6\%$ of instances, while propositional HER achieves $82.4\%$.
In comparison, state HER solves only $51.0\%$.

Table~\ref{table:coverage:lama} shows the performance of the baseline planner, LAMA.
Notably, the learned models often produce significantly shorter solutions than LAMA, indicating that they do not rely on random exploration.
Moreover, the learned models solve more instances than LAMA ($82.4\%$ vs.\ $80.8\%$), despite being restricted to a greedy policy without backtracking, whereas LAMA is allowed to backtrack.
Nevertheless, it is clear that the learned models learn general policies that can solve much larger instances, as shown in Table~\ref{tab:data}.

\subsection{Failures}

There are several domains where learning a general policy failed.
In \textbf{Delivery}, all packages must be delivered to the same location, but the truck can carry only one package at a time.
Random exploration can occasionally deliver a single package, but the likelihood of delivering multiple packages to the same location in a single trajectory is extremely low.
As a result, the models typically learn to deliver just one package, failing to generalize to the full task.

In the \textbf{Hiking} domain, a similar issue arises: the probability of generating informative trajectories through random exploration is simply too low for effective learning.

For \textbf{Childsnack}, and to some extent \textbf{Spanner}, learning a general policy is problematic.
Although coverage is high, this is due to selecting the best model across runs using the validation set.
The issue appears to be the handling of dead-end states: every trajectory ending in a dead-end is relabeled as a success for the relabeled goal.
In Childsnack, dead-ends occur when gluten-free ingredients run out; since some children have been fed, the state matches a subgoal.
Similarly, in Spanner, if some nuts have been tightened, the trajectory is relabeled as a success.
Consequently, there is no training data on how to avoid dead-ends for the original goal.

In \textbf{Reward} and \textbf{Visitall}, coverage decreases as the grid size increases.
This may be due to the model's depth: with larger grids, distant cells may not communicate effectively, limiting the model's ability to tackle larger instances.

\subsection{Training Curves}

Figure~\ref{fig:plots:blocks} shows the number of episodes required by each method to learn a model that achieves a given coverage on the test set of the Blocks domain.
Lifted HER learns models that reach nearly $100\%$ coverage with fewer episodes than propositional HER, and exhibits lower variance, especially in later episodes.
In contrast, state HER requires substantially more episodes to achieve comparable coverage, and its variance remains high throughout training.

We note that lifted HER and propositional HER have the ability to automatically generate a curriculum of increasing difficulty.
Figure~\ref{fig:plots:blocks} illustrates this by plotting the mean goal size and trajectory length during training.
Both lifted HER and propositional HER show a clear increase in goal size and trajectory length over time, indicating that the models are exposed to progressively harder tasks.
For lifted HER, the mean goal size initially decreases as it learns to deconstruct towers, but later increases as the model learns to solve the original task.
State HER is partially omitted from this figure, as its goal sizes remain large and relatively constant throughout training (varying with instance size).
However, state HER shows only a modest increase in mean trajectory length, indicating limited curriculum learning.

Due to space constraints, figures for the remaining domains are omitted; however, this is not limited to Blocks.

Automatic curriculum learning is not a new concept \cite{portelas-et-al-ijcai2020}, although it often requires a specific mechanism to implement it.
In our method, the curriculum emerges naturally through the relabeling process, which focuses on the largest subgoals that can be achieved at any given time.
While this is partially present in state HER and in the original formulation of HER, it becomes significantly more evident in propositional and lifted HER.

\subsection{Propositional vs. Lifted HER}

A noteworthy limitation of propositional HER is its reliance on achieving smaller subsets of the original goal.
This limitation appears in the Maze domain, where the goal is to reach a specific location.
The shortest path to the goal in the training set is approximately $65$ steps, but with a horizon of $100$ steps and the possibility of backtracking or taking dead-end branches, random exploration is unlikely to succeed.
Indeed, after running propositional HER for $120$ hours (across $10$ runs), not a single successful trajectory was found.

In contrast, lifted HER relabels the goal to the actual location reached by the agent.
This allows the model to learn a general policy, as demonstrated in Figure~\ref{fig:coverage:maze}, where it consistently achieves $100\%$ coverage.
State HER can relabel the goal, but these goals include static atoms, which appear to hinder learning; as a result, coverage never reaches $100\%$.

Propositional HER also has difficulty with tasks such as picking up a particular item among countless others, or clearing a specific block in Blocks when there are hundreds of blocks.
The probability of randomly generating a trajectory that achieves the precise goal is extremely low.
Lifted HER overcomes this by relabeling which object or block the agent was supposed to target, enabling it to start learning.

\section{Conclusions}

In this paper, we introduced three variants of Hindsight Experience Replay (HER) for generalized planning: state HER, propositional HER, and lifted HER.
These variants adapt HER to the planning setting, where states and goals are represented using first-order languages.
State HER relabels goals as full states, propositional HER restricts relabeled goals to subsets of ground atoms that appear in the original problem goal, and lifted HER uses lifted versions of these propositional goals that preserve suitable structural dependencies.
We showed that these HER variants can significantly improve the data efficiency and scalability of RL approaches to generalized planning.
In particular, they enable general policies to be learned in domains where it is not possible to precompute optimal values for all states.
We also demonstrated that these HER variants can automatically construct a more diverse and effective curriculum for learning, leading to faster convergence and better performance.

\section*{Acknowledgements}
{
    The research has been supported by the Alexander von Humboldt Foundation with funds from the Federal Ministry for Education and Research, Germany, by the European Research Council (ERC), Grant agreement No. 885107, and by the Excellence Strategy of the Federal Government and the NRW L\"{a}nder, Germany.
}
\bibliography{bibliography,crossref}

\end{document}